\documentclass{article}

\usepackage{arxiv}

\usepackage[utf8]{inputenc} 
\usepackage[T1]{fontenc}    
\usepackage{hyperref}       
\usepackage{url}            
\usepackage{booktabs}       
\usepackage{amsfonts}       
\usepackage{nicefrac}       
\usepackage{microtype}      
\usepackage{lipsum}
\usepackage{graphicx}
\usepackage{multirow}
\usepackage{marvosym}
\usepackage{float}
\graphicspath{ {./images/} }

\title{Retrieval-based Text Selection for Addressing Class-Imbalanced Data in Classification}

\author{
 Sareh Ahmadi \\
  Department of Computer Science\\
  Virginia Tech\\
  Blacksburg, VA 24061 \\
  \texttt{saraahmadi@vt.edu} \\
   \And
 Aditya Shah \\
  Department of Computer Science\\
  Virginia Tech\\
  Blacksburg, VA 24061 \\
  \texttt{aditya31@vt.edu} \\
  \And
 Edward A. Fox \\
  Department of Computer Science\\
  Virginia Tech\\
  Blacksburg, VA 24061 \\
  \texttt{fox@vt.edu} \\
}

\begin{document}
\maketitle
\begin{abstract}
This paper addresses the problem of selection of a set of texts for annotation in text classification using retrieval methods  when there are limits on the number of annotations (due to constraints on human resources). An additional challenge addressed is dealing with binary categories that have a small number of positive instances, reflecting severe class imbalance. In our situation, where annotation occurs over a long time period, the selection of texts to be annotated can be made in batches, with previous annotations guiding the choice of the next set. To address these challenges, the paper proposes leveraging SHAP  to construct a quality set of queries for Elasticsearch and semantic search, to try to identify optimal sets of texts for annotation that will help with class imbalance.
The approach is tested on sets of ``cue'' texts describing possible future events, constructed by participants involved in studies aimed to help with the management of obesity and diabetes. 
We introduce an effective method
for selecting a small set of texts for annotation and building high-quality
classifiers. We integrate vector search, semantic search, and machine learning classifiers to yield a good solution. Our experiments demonstrate improved F1 scores for the minority classes in binary classification.
\end{abstract}


\section{Introduction}
Annotation is an important step in the development and training of supervised machine learning models. It involves the process of adding labels to unlabeled data so algorithms will enable machine learning models to learn more effectively from that enhanced data. Annotations allow machine learning models to recognize patterns in a supervised fashion and make accurate predictions \cite{pustejovsky2012natural}.

 Therefore, annotation quality is a critical aspect of machine learning that can significantly impact the performance of a model. The quality of annotations refers to how accurate, consistent, and reliable are these labels, and how well the set of annotations provides sufficient and suitable examples. High-quality annotations are essential for machine learning models to achieve optimal performance. If the annotations are incorrect, inconsistent, or biased, the model's predictions can be unreliable, leading to poor performance. Therefore, it is crucial to ensure the quality of annotations before using them to train a model.
 
One way to ensure annotation quality is to have multiple annotators label each data point independently and then compare their annotations. That comparison can yield a metric, known as an inter-annotator agreement, that can help identify discrepancies between the annotations and provide insights into where the labels need improvement.

Another way to improve the quality of machine learning is to avoid having an unbalanced dataset, i.e., where the distribution of classes in a dataset is not equal. For example, in binary classification, one class could have a much larger number of samples than the other class. This can cause problems for machine learning algorithms as they may be biased toward the majority class and fail to identify the minority class. For example, if the majority class has more samples, the algorithm may become biased towards it and perform poorly on the minority class. Overall, dealing with unbalanced data requires careful consideration of the dataset and problem at hand.
The appropriate solution often will vary, depending on the specific circumstances.
Thus, one way to avoid the unbalanced scenario is by using an intelligent sampling method that aims to select data items for the annotation process that is likely to lead to a balanced distribution among the classes in the training set. This can help us to receive high-quality data for classification and build accurate classifiers. 

Overall, ensuring data quality is critical for building machine learning pipelines and it  can significantly impact the performance of the model. To help address this challenge, this paper aims to leverage retrieval models to avoid obtaining imbalanced training data.
One part of our solution involves performing a search on the pool of unlabeled data to identify positive examples by retrieval models.
Given that the annotation process involves large numbers of humans, and can take many weeks, there can be repeated use of search methods to help identify what data will be most informative if annotated.
Thus, multiple batches of data to be annotated can be identified, with the process of selecting a batch improving as annotations continue to arrive.
THus, it is possible to improve efficiency, quality control, and flexibility in the labeling process.
In particular, after receiving a batch, we can select the next one from the unlabeled data  considering the data distribution and  classification performance.
We can select the texts that can be more likely to make the classifier gain better performance. This can lead to  making the annotation effort more efficient.

Our proposed approach is tested on a dataset consisting of episodic future thinking (EFT) \cite{atance2001episodic} texts that are generated by people who suffer from different medical conditions such as diabetes or obesity. Participants are asked to write short stories for different time frames based on some given instructions.
\section{Literature review}
Search is a fundamental task in information retrieval (IR), which involves finding relevant information from a large collection of documents in response to a user's query. The goal of search is to retrieve the most relevant documents that satisfy the user's information needs.
In recent years, there has been a growing interest in using deep learning techniques for search in information retrieval \cite{bruch2022reneuir}. Deep learning models can learn to represent documents and queries in a dense high-dimensional space, where the similarity between documents and queries can be computed using various distance metrics, such as Euclidean distance, cosine similarity, or dot product. In this section, we review some methods for searching in information retrieval. 
\subsection{Semantic Search}

Semantic search is an information retrieval (IR) approach that aims to improve the relevance and accuracy of search results by understanding the meaning and context of queries and documents, rather than just relying on keyword matching \cite{bast2016semantic}. In semantic search, the system tries to identify the underlying concepts and relationships between terms, and use this knowledge to retrieve relevant documents, even if they do not contain the exact search terms. Two types of semantic search can be distinguished according to query length. There is a symmetric semantic search where the query and the passage have the same length.
The asymmetric case is where the query is shorter, e.g., a question or some keywords. The idea of semantic search is to embed all the documents as well as the query, and then perform a search in a vector space to find the most relevant documents relative to the query.

 Since the introduction of BERT \cite{kenton2019bert}, a powerful language model based on transformer neural networks, which can be fine-tuned for various natural language processing (NLP) tasks,  there has been growing interest in using deep learning models for semantic search.  BERT can produce dense vectors representing text (embeddings) for each word (or token), which can be used for semantic textual similarity (STS), semantic search, and clustering. One of the seminal papers on semantic search 
 \cite{karpukhin2020dense} proposes a dense vector-based approach to IR that is particularly well-suited for open-domain question-answering tasks leveraging BERT pre-trained models and a dual-encoder structure.

One approach to get embeddings from the texts and measure the similarity is to use a cross-encoder \cite{kenton2019bert}. The input to this model is two sentences separated by the SEP token.
They are passed to the model simultaneously. The model is followed by a classification head to predict a similarity score for the two sentences. The sentence embedding can be extracted by taking the average of all the token embeddings in the sentences or using the CLS token. While the performance of the cross-encoder is good, it requires computing embeddings and similarity for every sentence pair in the corpus, which is not efficient for big data.

A better approach uses Sentence-BERT
(SBERT) \cite{reimers2019sentence},
which can produce sentence embeddings much faster and is more scalable compared to a cross-encoder \cite{thakur-etal-2021-augmented}. SBERT is a bi-encoder that uses a Siamese network for training. The sentences are passed individually to the model.
Siamese BERT outputs a mean pooled sentence embedding  for each sentence. The two embeddings are concatenated and fed into a feed-forward neural network (FFNN) to predict the similarity score. The original work with SBERT \cite{reimers2019sentence} used natural language inference (NLI) datasets and softmax loss for training. Further work has been done using other language models such as  MPNet \cite{song2020mpnet} and RoBERTa \cite{liu2019roberta}, as well as multi-lingual models \cite{reimers2020making},\cite{chidambaram2019learning},\cite{yang2020multilingual}.
These have been trained on different datasets using the bi-encoder structure for generating sentence-embeddings.
Like with BERT, the embeddings from these models can be used across a range of applications.

Overall, semantic search is a rapidly evolving field that is driving innovation in IR and NLP. There have been several existing neural IR models \cite{thakurbeir} that can be leveraged to perform semantic searche and retrieve related documents. The techniques and algorithms developed so far are already making a significant impact on how we search for and extract information from large and complex data sources. 
Accordingly, one of the IR approaches we employ to improve the training datasets supporting text classification
is to utilize pre-trained models during the annotation process such that we obtain high-quality data to build powerful machine-learning models.

\subsection{Elasticsearch}

Elasticsearch \cite{elasticsearch} is a popular search and analytics engine used for indexing, searching, and analyzing large volumes of data. It provides powerful features for text-based search and retrieval, such as stemming, approximate matching, and relevance ranking. Elasticsearch can be combined with machine learning algorithms to improve search accuracy and relevance \cite{prasad2020enhancement}. Lu et al. \cite{lu2020auto} presented an auto-tuning method to improve the performance of Elasticsearch by using machine learning algorithms like random forest and gradient boosting regression trees. Kim et al. \cite{kim2022suggestion} propose a method to improve the classification and search of academic research results using the Elasticsearch classification method and the LDA-based topic modeling technique. Several frameworks use Elasticsearch to index and retrieve data, and machine learning algorithms to analyze user queries and provide personalized search results \cite{dong2022table, zamfir2019systems, liu2021document}. In the case of a large unlabelled real-world dataset, often we can face class imbalance issues where some common categories are more likely to be recorded by the participants \cite{johnson2019survey, liu2008exploratory, japkowicz2002class, japkowicz2000class}. While there has been a significant amount of research and development done on Elasticsearch, there is still a gap in exploring the combination of different search methods, so as to fully leverage them to aid downstream machine learning classifiers. More research is needed in this area to investigate how Elasticsearch can work in tandem with other search techniques to improve the accuracy and effectiveness of machine learning models in various applications. 

In this paper, we combine Elasticsearch with semantic search to identify texts from the pool of unlabelled data. The retrieved texts from these search methods can then be passed on to the annotators. In this way, we increase the probability of obtaining positively labeled cues for specific imbalanced categories, thereby resulting in a more balanced training dataset. That better balanced dataset can then be used to create more accurate classifiers for downstream tasks.

\section{Dataset}

Episodic Future Thinking (EFT) can be thought of as a scalable intervention to reduce Delay Discounting (DD) and improve health-related behaviors \cite{stein2016unstuck, stein2018episodic}.  Here, participants identify several events that may occur at multiple future time frames (e.g., 1 month to 10 years) and generate text-based event descriptions or \textit{cues} that can prompt the EFT. They generate vivid episodic descriptions of these events (cues), by completing an experimenter-guided interview or self-administered survey task. The data was originally compiled from 18 studies conducted by the medical research teams at 
two universities
that experimentally examined the effects of EFT on diabetes and other relevant health behavior outcomes. Adding the results from all the participants and studies, the overall dataset comprises an estimated 11,000 cues, each having an approximate length of a few sentences. Once the data is obtained, then manual annotation is used to label these cues. The annotators are from Amazon Mechanical Turk \cite{mturk_website}, a crowdsourcing platform where humans can complete intelligent tasks in exchange for monetary compensation. After every cue has been rated three times, the final ground truth is computed by aggregating all three ratings. In the case of binary categories, the final label is determined using a majority voting scheme applied to the three ratings. The data annotation is conducted in sequential rounds where a batch of data is passed at a given time to annotators. Once this data is annotated, the next set of data is passed and this process continues until collectively we have all of the required labeled data.  
Since funds for annotation are limited, our goal is to find the best 3000 cues wherein the annotations we receive will be as close as possible to the optimal training set for all 11 of our binary classifiers.

These cues are \textit{not} mutually exclusive; a given cue can belong to more than one of our 11 categories. Consider the following example: \newline

\texttt {In about 5 years, I am having a child. My sister and close friends wait with me in the hospital. It's my first child. We're nervous but is excited to welcome a new, healthy life into the world.} \newline

 The above cue would belong to two categories:  \textit{health} because it mentions a healthy life being brought to this world; and \textit{friends} because it mentions about close friends.

\begin{table}
\caption{Data Content Characteristics}
\centering
\def\arraystretch{1.5}%
\scalebox{0.8}{
\resizebox{\columnwidth}{!}{%
\begin{tabular}{ | c | c | p{6.5cm} | c | c | } 
\toprule
\textbf{Categories}  & \textbf{Label}  & \textbf{Definitions}  & \textbf{Samples} & \textbf{Total}  \\
\midrule
\midrule

\multirow{5}{*}{Alone}  & No  & Contains no references or mentions of any activities which portray loneliness or alone. & 1274 & \multirow{4}{*}{1554} 	\\ \cline{2-4}
                         & Yes & Contains an obvious, specific reference to events and activities which shows that they are being done alone.  & 280 &     \\ 
                                              
\midrule
\midrule

\multirow{4}{*}{Friends}  & No  & Contains no references or mentions of any friends. & 971 & \multirow{3}{*}{1554} 	\\ \cline{2-4}
                         & Yes & Contains an obvious, specific reference to friends discussing some event or activities. & 583 & \\
\midrule
\midrule

\multirow{6}{*}{Health}  &  No  & Contains no references to health; does not discuss physical state, mental health, or intentional changes in behavior and health outcomes.  & 1126 & \multirow{5}{*}{1554} \\ 
\cline{2-4}

 & Yes & Contains an obvious reference to physical or mental health. Examples may  describe mental and physical health, changes in behaviors, etc. & 428 & \\  
                                            
\bottomrule
\end{tabular}%
}}
\label{data content}
\end{table}


This paper considers a total of 3 different content categories from this dataset which possess a binary rating with severe imbalance. Table \ref{data content} illustrates the categories, the definitions given to annotators as explanation, and sample counts for each label obtained from a pilot study.
That study yielded $\approx$ 1600 labelled cues, with relatively few samples from the positive class in each of the 3 categories.

    \section{Methodology}

\subsection{Overview}

The purpose of this study is to build classifiers that predict a correct categorization
of the content of the EFTs. We are focused on the 3 different classes
previously described.
The goal is to label the most helpful subset of the data
to use in training.
After receiving the first batch
of the labeled data, data cleaning is performed.
To address the imbalance situations shown 
in Table \ref{data content}
we leverage retrieval models, e.g., we apply Elasticsearch as well as  semantic
search on unlabeled data
and retrieve cues likely to be related to those minority
classes. The retrieved data are given to annotators to be labeled to help reduce
data imbalance.
Given the first batch of imbalanced labeled data, we trained binary classifiers for each category using transformer models including BERT, XLNET \cite{yang2019xlnet}, and
 distiBERT \cite{sanh2019distilbert}, as well as an SVM classifer.
Since the data set is imbalanced, the macro F1-score needs to be reported for evaluation.

\subsection{Search for selection of texts}

 For semantic search, an existing pre-trained sentence embedding model is needed to embed the  texts in the unlabeled dataset and create a searchable vector database. The model embeds each cue (text) into a vector space such that it is the semantic representation of that specific text.  There are several  different pre-trained  models for embedding search queries and the texts to perform a semantic search. Based on  the sentence-BERT library,  all-mpnet-base-v2 has been shown to have the best performance on semantic search \cite{Jones1}. The base model is MPNet \cite{song2020mpnet} and it is trained on a very large and diverse  dataset of over 1 billion training pairs. The mean pooling operation is done to generate the embeddings for each text in the dataset.
 The model produces normalized embeddings of size 768 for each text in the unlabeled dataset.
 The neural retriever model encodes the query into 768 dimension vectors, and then computes a similarity score  between the query vector and all text vectors.
 We consider Euclidean distance   for the similarity since it is more accurate compared to other metrics.

Semantic search can be done with vector databases to store and index vector embeddings from language models. There are different libraries to index and perform searches. Facebook AI Similarity Search (Faiss) \cite{johnson2019billion} is one of the most popular implementations of efficient similarity search. This library can be used to retrieve texts related to the corresponding query in a fast and efficient way. It includes GPU support that can be leveraged  at inference time to perform searches on a huge dataset.

Besides performing a semantic search by using a neural retrieval model, we have also leveraged the Elaticsearch tool for searching. For both models, we used the same queries for each of the three categories. 

\begin{figure}[h!]
\centering
\includegraphics[width = 0.7\linewidth]{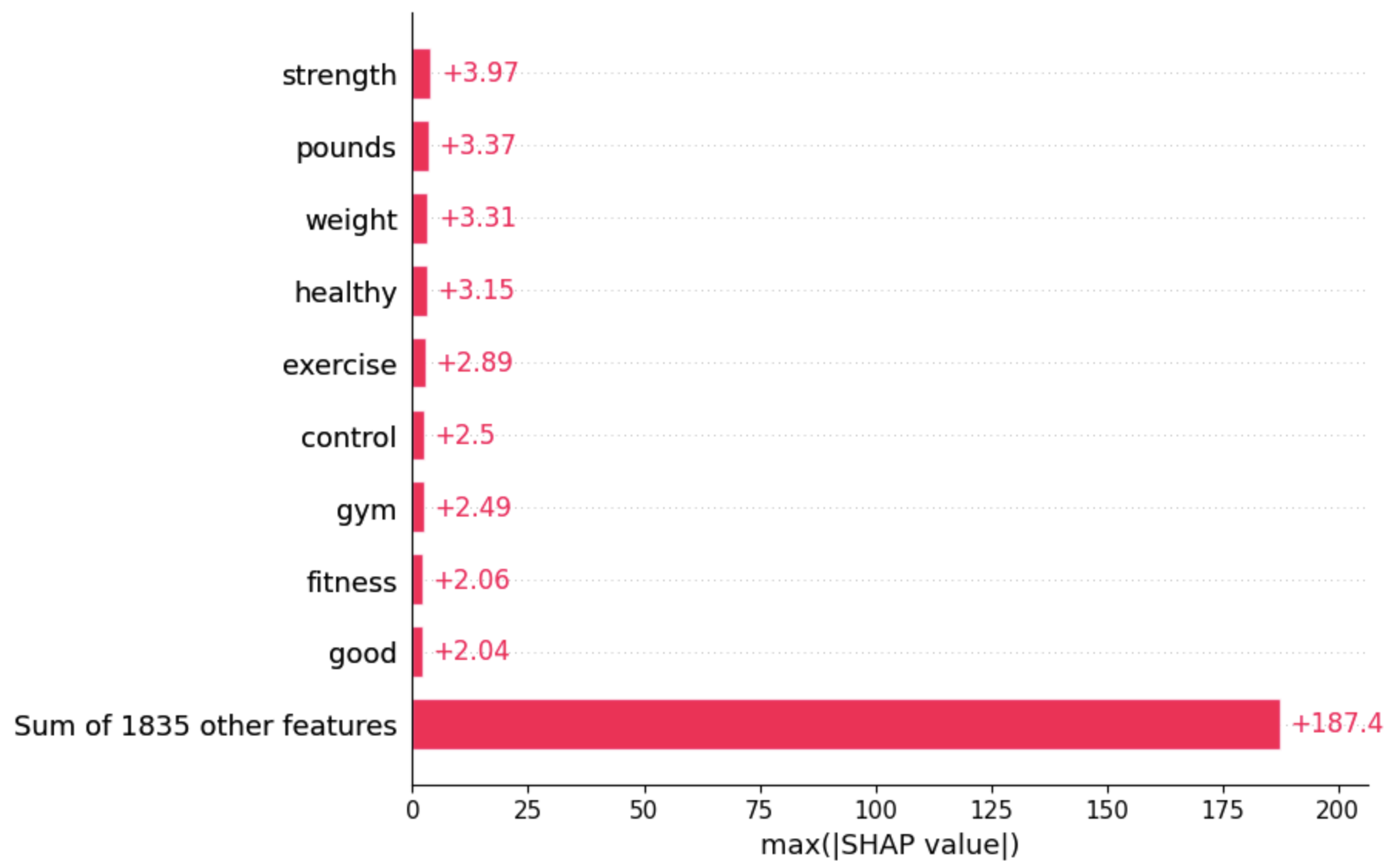}
\caption{Top words obtained using SHAP for classifier trained on ``Health'' category}
\label{fig:5-1}
\end{figure}

\begin{figure*}[t]
\centering
\includegraphics[width = 15cm]{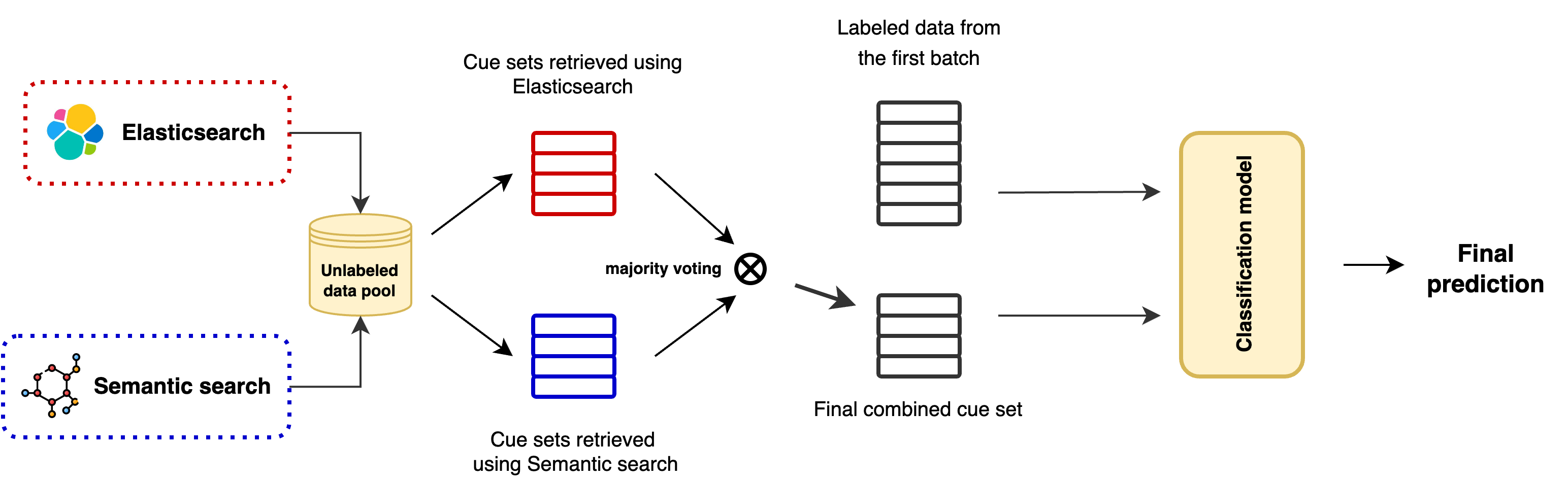}
\caption{Overview of the proposed framework. Elasticsearch and Semantic search is executed on the unlabeled pool of data using the same set of queries to obtain positive samples for each category. The resulting cue sets obtained from both methods are combined with the previously labeled data and used for modeling.}
\label{model}
\end{figure*}

To identify potentially useful cues from the pool of unlabelled data, tailored queries are used for every category. These queries are usually keyword-based, and Elasticsearch uses them to find cues that approximately match these keywords. To create these queries, we leverage the classifiers which are initially trained on the first batch of labeled data shown in Table \ref{data content}. Once we have the results from trained classifiers, we utilize SHAPley Additive exPlanations (SHAP) \cite{10.5555/3295222.3295230}, a popular method for studying model predictions. SHAP values enable us to visualize and understand the important parts of the input data that the model focuses on while predicting the output label. SHAP values provide both global and local explainability, which helps us understand how the features of the input data are related to the outputs. By using SHAP on the trained classifiers, we aim toidentify an optimal set of keywords that the model focuses on while predicting the cue as positive. For example, Figure \ref{fig:5-1} shows the top words generated by SHAP for the ``health'' category.

We manually verify these words to form the best set of words, which can be used as a query to retrieve  cue sets from both Elasticsearch and semantic search. We follow this procedure for each of the categories to obtain queries for those categories.

Table \ref{queries} shows different queries that are helpful to retrieve cues for
the positive class of a category.

\begin{table}[ht!]
\centering
\caption{Queries used for categories}
\def\arraystretch{1.4}%
\begin{tabular}{ c | c} 
\toprule

\textbf{Categories}  &  \textbf{Queries}  \\
\midrule

Alone & \texttt{alone  myself} \\
\hline

Friends & \texttt{friend} \\
\hline

Health & \texttt{health weight gym exercise} \\

\bottomrule
\end{tabular}
\label{queries}
\end{table}


As shown in Table \ref{queries}, different categories might have a set of specific keywords which capture a positive class context within the cue. For example, for a cue to be positive or highly related to the \textit{health} category, it generally will have some mention of words that indicate health like ``exercise'', ``gym'', and``healthy.'' When the cues are searched using a query, Elasticsearch will retrieve the top-ranked cues, along with a similarity score assigned to each of them. 

Elasticsearch uses  BM25 (Best Matching 25) \cite{robertson2009probabilistic}, a probabilistic information retrieval model to rank documents based on their relevance to a given search query. It takes into account both the term frequency (how often a term appears in a document) and the inverse document frequency (how often a term appears in the entire collection of documents). In BM25, each query term is given a weight based on its frequency and importance to the query. The weight is then combined with the weights of other terms in the query to estimate a relevance score for each document. The scores are sorted in descending order to rank the documents.
The queries mentioned in Table \ref{queries} are used on the entire pool of unlabelled data to retrieve example cues for every required category.

Therefore for each text in the dataset, we obtain a similarity score between the query vector and the text vector. The
top k texts from both semantic search and Elasticsearch are extracted. To ensure the final identified text is relevant given that we have different search performance for different search methods and categories,
we consider the text that is present in both
top sets.
In other words, we consider the majority vote from both tools to select the final texts for annotation. This will help us to construct a set of cues for training high-quality classifiers. This set is then passed on for annotation, increasing the likelihood of them being labeled as positive when reviewed by annotators and thereby achieving a better balance in training datasets. After receiving the new batch of the data with newly identified positive examples, we trained the classifiers on the new data and measured the F1-score.
The results after identifying the positive examples by performing a search are reported in the Results section. The overview of the proposed method is depicted in Figure \ref{model}.

\subsection{Classifiers}

Our dataset consists of cues which are textual descriptions of events or plans. This work includes leveraging knowledge from transformer models by fine-tuning them on the EFT dataset to identify various categories within a cue. For binary content characteristics, the model is trained to predict whether a given cue belongs to a respective category. 

Consider an input text cue, $X = [x_1, ..., x_n]$, where $x_i$ is the $i-$th token in the text and $n$ is the length of the sequence. This sequence of tokens is passed as an input and uses the same tokenizer library as used by the respective transformer model. Since this is a classification task, we obtain the embedding for the $[CLS]$  token from the transformer model. Let $C \in R^d$ be the output from the classification head with hidden dimension size $d$. This output is passed to feed-forward networks followed by a softmax layer for final prediction.

BERT \cite{kenton2019bert}, is a pre-trained deep learning model based on the transformer architecture and is trained on a large corpus of text data to learn the context and meaning of words. BERT is bidirectional, meaning it can process both the left and right context of a word simultaneously, making it more effective at understanding the relationships between words in a sentence. DistilBERT\cite{sanh2019distilbert} is a smaller and faster version of BERT It has fewer layers and fewer parameters, making it more lightweight and efficient than the original BERT model. DistilBERT is designed to be used in applications where speed and resource efficiency are important. XLNet\cite{yang2019xlnet} is another pre-trained deep learning model based on the transformer architecture and uses a permutation language modeling approach, where it considers all possible permutations of the input sequence instead of just the forward and backward directions as in BERT. 

Besides deep learning models, we have also used the SVM classifier. Support Vector Machines (SVMs) are a popular machine learning algorithm that can be used for text classification. SVMs work by finding a hyperplane that separates the data into different classes, with the largest margin possible between the classes. In the case of text classification, the SVM algorithm learns a model that maps documents to a set of features, which can then be used to classify new documents.

\begin{table*}
\centering
\caption{Performance on all the datasets. Each experiment is run for 3 trials and the average result is reported.\hspace{1mm} \textbf{*} \hspace{1mm} indicates the most recent cue sets for the minority class obtained using both the search methods. \textbf{bold} indicates the best results obtained. }
\def\arraystretch{1.4}%
\scalebox{0.59}{
\begin{tabular}{ c | c | c | c | c | c | c | c | c | c }
\toprule
\multirow{2}{*}{\textbf{Category}} & \multirow{2}{*}{\textbf{Data Samples}} & \multicolumn{2}{c|}{\textbf{SVM}} & \multicolumn{2}{c|}{\textbf{DistillBERT}} & \multicolumn{2}{c|}{\textbf{BERT}} & \multicolumn{2}{c}{\textbf{XLNet}} \\ 

\cline{3-10}
&  &  \textbf{Minority F1 score} & \textbf{Macro F1 score}  & \textbf{Minority F1 score} & \textbf{Macro F1 score}  &  \textbf{Minority F1 score} & \textbf{Macro F1 score}  &  \textbf{Minority F1 score} & \textbf{Macro F1 score} \\
\midrule
\midrule

\multirow{3}{*}{Alone} & 1600  & 0.66 & 0.79 & \textbf{0.87} & \textbf{0.92}  & 0.87 & 0.92 & 0.76 & 0.86    \\
\cline{2-10}

                    & 1600 + 25  & 0.65 & 0.79 & 0.87 & 0.92  & 0.88 & 0.93 & 0.77 & 0.86    \\
\cline{2-10}

                       & 1600 + 69\textbf{*}  & \textbf{0.70} & \textbf{0.82} & 0.89 & 0.93 & \textbf{0.90} & \textbf{0.96} &\textbf{ 0.79} & \textbf{0.88}  \\                                      
\midrule
\midrule

\multirow{3}{*}{Friends}  & 1600  & 0.74 & 0.85 & 0.80 & 0.89  & 0.82 & 0.90 & 0.84 & 0.90    \\
\cline{2-10}
                    & 1600 + 25  & 0.73 & 0.84 & 0.85 & 0.91  & 0.85 & 0.91 & 0.84 & 0.91    \\
\cline{2-10}
                       & 1600 + 55\textbf{*}   & \textbf{0.76} &\textbf{ 0.86} & \textbf{0.91} & \textbf{0.95}  & \textbf{0.89} & \textbf{0.94} & \textbf{0.87} & \textbf{0.92} \\
                                              
\midrule
\midrule

\multirow{3}{*}{Health} & 1600   & 0.77 & 0.85 & 0.88 & 0.91  & 0.88 & 0.91 & 0.91 & 0.94   \\
\cline{2-10}
                    & 1600 + 35  & 0.83 & 0.87 & 0.88 & 0.91  & 0.89 & 0.93 & 0.92 & 0.94    \\
\cline{2-10}
                       & 1600 + 87\textbf{*}   & \textbf{0.85} & \textbf{0.88} & \textbf{0.90} & \textbf{0.93} & \textbf{0.90 } & \textbf{0.93} & \textbf{0.93 } & \textbf{ 0.95 }\\
\midrule                                       
\bottomrule
    \end{tabular}}
    \label{result}
\end{table*}


\subsection{Experiment details}

This research includes experiments with three transformer models: BERT, its distilled  version DistillBERT, and XLNet. The used versions are \textit{bert-base-cased}, \textit{distilbert-base-cased}, and\textit{xlnet-base-cased}, respectively, all  from Hugging Face \cite{huggingface}. For  XLNet, BERT, and DistillBERT, we get an embedding dimension of 768 ($d  = 768$). The model is trained for 10 epochs with a learning rate of $2e-5$ and the Adam optimizer. The batch size is 16 for all of the experiments. The experiments follow a \textit{train : val : test} split of 70\% : 10\% : 20 \%. The model is fine-tuned on training data and evaluated on a validation set. All of the experiments are performed on NVIDIA's A100 GPUs. During training, after every epoch, the F1 score is calculated on training and validation data. If the current F1 score obtained on validation data is better (higher) than before, then that better model checkpoint is saved. All of the experiments are conducted for 3 trials, and the average result is reported. During testing, the checkpoint for the best model is loaded and evaluated on test data.


\begin{table}[ht!]
\centering
\caption{Precision and Recall for search methods}
\def\arraystretch{1.4}%
\scalebox{0.8}{
\begin{tabular}{ c | c | c | c | c | c} 
\toprule

\multirow{2}{*}{\textbf{Category}} & \multirow{2}{*}{\textbf{Data Samples}} & \multicolumn{2}{c|}{\textbf{Elasticsearch}} & \multicolumn{2}{c}{\textbf{Semantic search}}\\

\cline{3-6}
&  &  \textbf{Precision} & \textbf{Recall}  & \textbf{Precision} & \textbf{Recall}\\
\midrule
\midrule

\multirow{4}{*}{Alone} & Top 20 &  0.9  & 1.0 &  0.8  & 1.0 \\
\cline{2-6}
                        & Top 40 &  0.86  & 1.0 &  0.6  & 1.0 \\
\cline{2-6}
                        & Top 60 &  0.85  & 1.0 &  0.51 & 1.0 \\
\cline{2-6}
                        & Top 80 &  0.85  & 1.0 &  0.47  & 1.0 \\
\midrule

\multirow{4}{*}{Friends} & Top 20 &  0.85  & 1.0 &  1.0  & 1.0 \\
\cline{2-6}
                        & Top 40 &  0.86  & 1.0 &  0.92  & 1.0 \\
\cline{2-6}
                        & Top 60 &  0.92  & 1.0 &  0.8  & 1.0 \\
\cline{2-6}
                        & Top 80 &  0.89  & 1.0 &  0.7  & 1.0 \\
\midrule

\multirow{4}{*}{Health} & Top 20 &  1.0  & 1.0 &  1.0  & 1.0 \\
\cline{2-6}
                        & Top 40 &  0.98  & 1.0 &  1.0  & 1.0 \\
\cline{2-6}
                        & Top 60 &  0.97  & 1.0 &  1.0  & 1.0 \\
\cline{2-6}
                        & Top 80 &  0.98  & 1.0 &  1.0  & 1.0 \\
\midrule

\bottomrule
\end{tabular}}
\label{query performance}
\end{table}


\section{Results}
To evaluate the performance of the semantic search and Elasticsearch, we perform a search on the first batch of labeled data consisting of 1600 labeled cues.
We retrieved data for each category considering the top 20,40,60,80 cues as positive examples (label 1) based on their similarity score. We measure the precision and recall for the retrieved data given that we have the true label for each retrieved example.  The results are shown in Table \ref{query performance}.

Given that we considered all the retrieved examples as positive, the recall is 1 for all the categories, since there are no false positives. For the semantic search, the best performance comes from the health category with a precision of 1 for all the top k cues, i.e., given the query, everything it returns belongs to the health category (true label). The performance for the friend category for the top 60 cues is  more than 80\%; for the alone category the top 40 cues have precision over 60\%.

For Elasticsearch,  similar to semantic search, the health category has the best search results.
But the performance of semantic search is better for this category. However, the search result for the alone category in Elasticsearch is much better compared with semantic search, with a precision of 85\% for the top 80 search results. For the friend category, semantic search surpasses Elasticsearch for the top 40 retrieved data, but after that, Elasticsearch has better performance.

The results in Table \ref{query performance} show different search performance for the three different categories if we consider the same query. For this reason, we consider the majority vote for each retrieval model to ensure that the retrieved data is present in both  models and it is likely to be a positive sample for that specific category.

Table \ref{result} shows the performance of the classifiers for the three categories. The first row for each category is the performance when there is the original 1600 labeled data. It is seen that the F1-score on the minority class is less than 90\% for all of the models. The second row is after identifying a small sample for each  category, i.e., 25 positives for the alone categoryand the friend category, and 35 for health. For this scenario, the performance of BERT and XLNET has increased 1\% for the alone category from 87\% to 88\% and 76\% to 77\%, for transformer BERT and XLNET, respectively.  For the friend category, the BERT model has improved the F1-score to up to 85\%. For the health category, transformers BERT and XLNET increase by 1\%, and distilBERT increases  from 88\% to 90\% . Given that the first round of search improved the F1-score, we performed another search round and identified more examples. For health, we identified 52 more positives, for alone 44, and for friend 33.
We ran the experiments on this newly labeled data. The results are the third row for each category. It is shown that all transformer models have gained more F1-score for alone, i.e., 2\% more.
The best transformer is BERT, with macro f1-score of 96\%.
For friend, the distilBERT F1-score is increased form 85\% to 91\%.
BERT has increased from 85\% to 89\% and XLNET from 84\% 87\%. Overall the improvement for all the transformer models is significant for this category.
For health, the improvement for distillBERT is 2\%, from 88 to 90, and BERT and XlNET increase 1\% to 90\% and 93\%, respectively
The improvement for the SVM classifier is significant in all three categories, ranging from 4\% for the alone category, 2\% for friends, and 8\% for the health category for the minority class.

\section{Conclusion}
In this research, we proposed a novel method to identify positive examples in unlabeled data for binary classification when there is a class-imbalanced dataset. The identification is done during the annotation process such that we can select data samples that are likely positives for the classification task. We utilized retrieval models including a neural retriever to perform a semantic search, and keyword search using the Elasticsearch tool.
We applied them to the pool of unlabeled data to retrieve data examples for the minority class in order to build machine learning models with better performance. We consider a sentence-bert model trained on a huge information retrieval dataset for the neural retrieval model, and BM25 in Elasticsearch. To ensure that the retrieved examples from both  retrieval models are related to the given query, i.e., are positive examples for the specific category, we consider the majority vote for two sentence-bert and BM25 methods.  Three different  transformer models including BERT, distillBERT, and XLNET, are trained as well as  SVM classifiers for the purpose of predicting the content of the episodic future thinking (EFT) dataset. It is observed that identifying the positive examples increased the F1-score for the minority class leading to more accurate classifiers. For the transformer models, the more data that is identified, the better performance can be gained. For the SVM algorithm the improvement is more significant, suggesting a very effective method for non-neural machine learning algorithms when they do not require a lot of data examples. For future work, an approach can be introduced for intelligent sampling to select optimal cues considering other constraints within the dataset. Fine-tuning of the neural retrieval models can be done if a large quantity of labeled data is available to gain better performance for the semantic search.









\bibliographystyle{unsrt}  

\bibliography{references.bib}




\end{document}